\documentclass[runningheads]{llncs}
\usepackage[T1]{fontenc}
\usepackage{graphicx}
\usepackage{tabularx,ragged2e}
\usepackage{booktabs}
\usepackage{float}
\usepackage{amsmath}
\begin{document}
\title{Collaborative Multiobjective Evolutionary Algorithms in the search of better Pareto Fronts. An application to trading systems}
\titlerunning{Collaborative MOEAs}
%
%\titlerunning{Abbreviated paper title}
% If the paper title is too long for the running head, you can set
% an abbreviated paper title here
%

\author{Francisco J. Soltero\inst{1} 
\and\\
Pablo Fernández
\and\\
J. Ignacio Hidalgo\inst{2}\orcidID{0000-0002-3046-6368} 
 } 
%
% First names are abbreviated in the running head.
% If there are more than two authors, 'et al.' is used.
%
\institute{Universidad Rey Juan Carlos \email{franciscojose.soltero@urjc.es}
\and
Universidad Complutense de Madrid, Spain \\ \email{hidalgo@ucm.es}}
\maketitle              % typeset the header of the contribution
\begin{abstract}
Technical indicators use graphic representations of data sets by applying various mathematical formulas to  financial  time series of prices. These formulas comprise a set of rules and parameters whose values are not necessarily known and depend on many factors: the market in which it operates, the size of the time window, and others. This paper focuses on the real-time optimization of the parameters applied for analyzing time series of data. In particular, we  optimize the parameters of technical financial indicators and propose other applications such as glucose time series. We propose the combination of several Multi-objective Evolutionary Algorithms (MOEAs). Unlike other approaches, this paper applies a set of different MOEAs, collaborating to construct a global Pareto Set of solutions. Solutions for financial problems seek high returns with minimal risk. The optimization process is continuous and occurs at the same frequency as the investment time interval. This technique permits to apply the non dominated solutions obtained with different MOEAs at the same time. Experimental results show that this technique increases the returns of the commonly used Buy \& Hold strategy and other multi-objective strategies, even for daily operations. 
\end{abstract}
\section{Introduction}
Time series have multiple applications nowadays in fields such as medical applications and the financial market. One of the most important financial markets is the Foreign Exchange (Forex). International trade sharing and exchange are made at Forex. In other words, Forex is a financial market in which the values of currencies are traded. Initially, this exchange was due to the sale of goods and services. However, today these operations represent a tiny percentage of its activity, and most transactions are due to operations related to the trading of financial products. Consequently, this market is independent of the variations in actual trade flows. It is more influenced by other variables, such as the growth of GDP, inflation, the mass of interest, and trade balance, among others.\\

The main activity of the Forex market is currencies exchange, and the value of a data in the time series (price) is the exchange rate between them. This relationship is called the nominal exchange rate and indicates the amount of currency the market provides for one monetary unit for the other. The expression more commonly used to express the exchange rate between two currencies is the ratio  Currency A / Currency B. For instance, USD/EUR represents the exchange value of one US dollar per Euro.\\

The behavior of any type of financial asset, including currencies, has been observed since the creation of this type of market \cite{chopra1992measuring,lehmann1990fads,lo1990contrarian,allen1999using}. The way of making this study can be  classified as fundamental or technical analysis. Fundamental analysis (FA) studies the price of an asset from the point of view of the future evolution of its performance \cite{dechow2001short}. Technical analysis (TA) \cite{murphy1999technical} is based on market action study, mainly through graphics. For both, it is essential to identify the tendency in the early stages so that the following operations are done in the right direction. Of course, the final objective of the analysis is to construct trading rules and follow them to operate in the market, making exchange operations to get economic returns with the minimum risk.\\

TA is based on technical stock market indicators (Tis) configured according to a set of parameters, which work on discrete time series of prices of the target value. There is a wide range of TIs, some simple, others more elaborate. All TA tools intend to obtain relevant information to help investors to make sound investment decisions, defining buying and selling operations, even under conditions of uncertainty.\\

There are different ways to evaluate the results obtained by these techniques. One of the most common is to compare the results obtained with the strategy of "Buy and Hold" (B \& H). This is based on the view that long-term financial markets get a reasonable rate of return despite periods of volatility or loss. Buy and Hold is one of the strategies most commonly used by investors, not only to compare but also for investing.
Several studies have applied Evolutionary Computation techniques to obtain trading rules to maximize the benefits of investments \cite{chen1997toward,allen1999using,chen2002genetic,malkiel2019random,sweeney1988some}. A good review of them was done by Lohpetch and Corne in \cite{lohpetch2009discovering,lohpetch2010outperforming,lohpetch2011multiobjective}, where the use of Multi-objective Genetic Programming is suggested to discover effective trading rules. More recently, Bodas et al. \cite{bodas2009multiobjective,fernandez2008technical} showed that parameter optimization of TIs with Multi-objective Evolutionary algorithms (MOEAs) could generate investment strategies that improve the performance of B \& H.\\

In this paper, we expand the above proposals with a new approach that uses different evolutionary algorithms to obtain the best values of the parameters of technical indicators and generate investment rules. The novelty is that we construct the Pareto set of solutions by combining those obtained with different MOEAs. In particular we apply NGSA-II \cite{deb2000fast}, SPEA-II \cite{zitzler2001spea2}, PAES \cite{knowles1999pareto}, PESA-II \cite{corne2000pareto} and MoCell \cite{nebro2009mocell}. These MOEAs are important references in the literature and were designed to cover several parts of the search space. We propose to make them collaborate in the exploration of the search space. For this aim, this paper compares two different approaches. The first approach (approach A) is based on a classical approach, where unique MOEAs are used to get all non-dominated solutions. In the second one (Approach B), the best solutions of the non-dominated solutions are selected from those obtained by the set of MOEAs indicated above. Experimental results show that this new technique improves strategies A and the Buy \& Hold approach. Furthermore, results are good for different time windows.\\

The main contributions of this work are:
\begin{itemize}
    \item We propose a new  evolutionary algorithm approach where different well-known algorithms collaborate to solve multi-objective problems.
    \item We apply this approach to the real world problem of generating trading rules for the Forex exchange market. 
    \item We compare with approaches that use only one MOEA algorithm and show that our proposal of combining different algorithms covers a more extended part of the Pareto front and obtains non dominated solutions.
\end{itemize}
The rest of the paper is structured as follows. Section 2 briefly reviews related work on Evolutionary Computation and finance. Section 3 explains the method proposed in this paper. Section 4 presents the time-real environment. Section 5 shows the experiments performed. Section 6 presents the results, and. the conclusions are presented in section 7.

\section{Related Work}
\label{sec:related_work}
Evolutionary computing has been widely applied to problems of prediction and optimization related to economics, finance, and, more specifically, the currency market. Both mono-objective and MOEA have been used in a variety of problems \cite{hassan2010multiobjective}: Portfolio Optimization and Stock Selection \cite{deb2000fast,lohpetch2009discovering}, Pricing Derivatives \cite{fan2007option}, Management of Financial Risk \cite{oussaidene1997parallel,schlottmann2005multiobjective}, Forecasting and Time Series Prediction \cite{butler2009multi,pavlidis2005computational,trojanowski1999evolutionary}, Evolving Technical Rules for Trading and Investment \cite{allen1999using,neely2003risk}, and Decision Making \cite{tsang2000eddie}.\\

In this article, we focus on applying MOEAs to optimize the parameters of technical indicators within the FOREX Market.  A wide variety of researchers seek to profit in this market using different techniques. Some approaches use Genetic Algorithms or other Evolutionary Algorithms to optimize a Neural Network \cite{hussain2016regularized,dorsey1998use,rout2014forecasting,yaman2014evolutionary}. Algorithms based on Genetic Programming (GP) \cite{stijven2014optimizing}. EA´s to evolve trading rules \cite{alexander1961price,chang2014pso,chen2002genetic,lohpetch2011multiobjective,parisi2006models}.  In \cite{fama1966filter} and \cite{bodas2009multiobjective} it was proposed, a version of a technique for optimizing the parameters of TIs such as the Moving Average Convergence-Divergence indicator (MACD) and the Relative Strength Index (RSI) \cite{fama1966filter}. The technique is based on using MOEAs with Super Individual (MOEASI) and can be applied many times.\\

The training data set, constraints, and objective function are static in dynamic optimization problems. Usually, the optimization starts from a fixed interval of data, called training sets, from which it obtains a group of solutions. These solutions undergo a validation period , where they pass various filters based on restrictions and new data sets. The set of resultant individuals will be used in the evaluation periods.\\

Using this static data set to obtain individuals generates mainly two problems: 
\begin{enumerate}
    \item It does not take into account the evolution of the system, its current status, and the constant changing. 
    \item Individuals may be sensitive to the training data set, producing the undesired effect of overfitting \cite{fama1966filter}.
\end{enumerate}
Validation periods named above are included to avoid this overfitting. Thus, the sensitivity to the data set decreases but is not eliminated. It would be possible to consider sensitivity to a more extensive data set.
Different authors have studied the characteristics of changes in dynamical systems \cite{branke2003designing,corne2000pareto,lohpetch2009discovering}. This paper uses the “tick by tick” series of the EUR/USD pair. The characterization of this series could be modeled on a system of small movements and moderate changes \cite{jong1999evolving}. The time series dynamic is vital because too-sharp dynamics could limit the provision of information to the next generation of solutions.\\

\section{Methodology}
\label{sec:materials}
This paper investigates two approaches for optimizing TIs based on MOEAs. Solutions to the optimization problem represent the parameters of the technical indicators selected for obtaining trading signals. This TI, with its parameters, will be used to decide what to do (buy - sell - nothing) at the beginning of the following day (or the subsequent time step). Unlike other approaches, in the proposals presented here, non-dominated solutions are obtained continuously and applied during a limited period. This makes them highly adaptable to market conditions. Moreover, these solutions provide information about the system's current state for future iterations, as they are part of the initial population for the next time intervals. 
The approach tackles the dynamism of the problem, since it considers the current state and changes of the system. The active individuals will be the best for the current data set. When new data enter the system, they will activate the computation of new solutions. The number of new solutions is controlled and is around 30 percent of the total population. 
\subsection{Two different combinations of MOEAS}

\begin{figure}[!h]
\label{fig:1}
\includegraphics[width=0.65\textwidth]{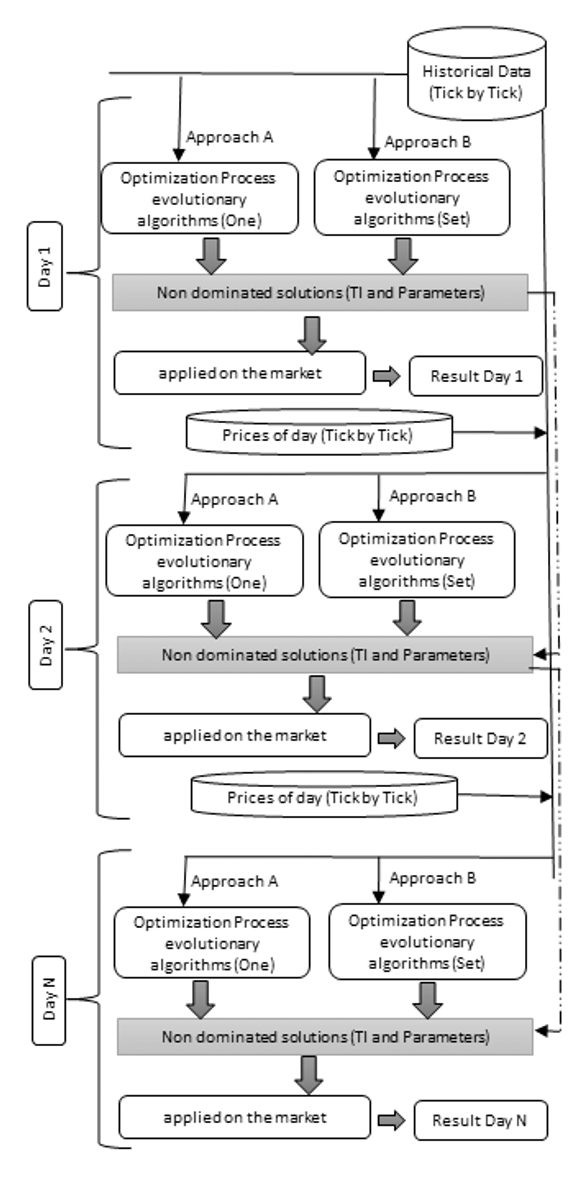}
\caption{Technique overview with approaches A and B}
\end{figure}

As we have mentioned, two different approaches are compared. They differ in the way the solutions are obtained: 
\begin{itemize}
    \item Approach A: Apply a single MOEA  for getting non-dominated solutions. 	
    \item Approach B: Combine a set of MOEAs to obtain a global group of non-dominated solutions. 
\end{itemize}
Figure \ref{fig:1} shows the functional diagram of both approaches. Under Approach A, a single MOEA is used, whereas under Approach B a set of them is used, but in both cases, the result of the optimization process is a set of non-dominated solutions. These are applied in the market for obtaining daily results. In both approaches, the solutions are selected according to a group of objectives under a non-dominating selection process. The fitness values are calculated by simulating the market with the selected data set.
Approach A allows us to investigate various implementations of different MOEAs, while Approach B constitutes the main contribution of this work. We have selected a set of the most popular MOEAs found in the literature: 
\begin{itemize}
    \item Non-dominated Sorting Genetic Algorithm-II (NSGA-II) \cite{deb2000fast}.
    \item Strength Pareto Evolutionary Algorithm (SPEA2) \cite{zitzler2001spea2}.
    \item Pareto Archived Evolution Strategy (PAES) \cite{knowles1999pareto}.
    \item Niched Pareto Genetic Algorithm (PESA-II) \cite{corne2000pareto}.
    \item Cellular Genetic Algorithm (MOCell) \cite{nebro2009mocell}.
\end{itemize}

%%%%%%%%%%%%%%%%%%%%%%%%%%%%%%%%%%%%%%%%
\subsection{Genetic Encoding}
\label{sec:genetic_encoding}
As in other evolutionary computation algorithms, solutions are represented by a set of integer chromosomes or a group of values. To obtain a trading rule, i.e., a solution, we need the parameters of two technical indicators, their corresponding time windows, the operational parameters, and the triggering signals. According to that, four chromosomes that give us the values of the needed different parameters compose the evolutionary algorithm code.

\subsubsection{Chromosome for the two technical indicators:}
We have selected the Moving Average of Convergence-Divergence indicator (MACD) and the Stochastic Oscillator indicator (SOI). 
MACD is obtained in a plot with two lines given by equation \ref{eq.1} and equation \ref{eq.2}: where EMA(a, Y(t)) indicates the exponential moving average over the period a over the time series Y(t). Therefore, three parameters are necessary to define the MACD indicator; a, b and c.
\begin{equation}
    \label{eq.1}
    \text{MACD}(t)=\text{EMA}(a, Y(t)) - \text{EMA}(b, Y(t)) 
\end{equation}

\begin{equation}
    \label{eq.2}
    \text{SIGN}(t)=\text{EMA}(c,\text{MACD}(t))
\end{equation}
On  the other hand, the SOI is a momentum indicator that uses support and resistance levels. The term “stochastic” refers to the point of a current price about its price range over a previous period. It attempts to predict value turning points by comparing the closing price to its price range. Two levels based on the history of the closing values delimit the price range. Traditionally the content is fixed between 20\% (overbuy) and 80\% (oversell) of the maximum value over a period; in this paper, dynamic bands adapt their position to the recent price behavior, replacing those levels. SOIa and SOIb delimit the band. In addition, we need another parameter to calculate the indicator, SOIc. This indicates the period to apply equations (3) and (4). \%K generates an anticipatory signal, and \% D is a signal confirming the trend in the change.
\begin{equation}
\label{eq3}
   \%K = 100 \cdot \frac{Price-Lowest Price(SOIc)}{Highest Price(SOIc)-Lowest Price(SOIc)}
\end{equation}
\begin{equation}
\label{eq4}
\%D = SMA_3 (\%K)	
\end{equation}
where $SMA_3$ is the 3-period simple moving average of \%K and Price is the last closing price.In summary, the first chromosome represents six parameters to be optimized: MACDa, MACDb, MACDc, SOIa, SOIb and SOIv.\\

\subsubsection{Chromosomes for the time windows:}
This is one of the most innovative aspects of this work because investors usually work with standard-size windows in a traditional trading system. In this paper, we let the algorithm select the optimal size of the window to operate with MACD and SOI. The solution strategy chooses the time window in which a higher yield is obtained. The parameters are ESTC1 and ECTS2 for the timescales of MACD and SOI, respectively. The value is measured in the number of ticks and can take a value bellowing to [1, 14400].\\

\subsubsection{Chromosome related to the operation of the market:}
The operational parameters are three: "stop-loss", "take-profit" and "trailing-stop". Those values are commonly used in trading and are measured in pips with a numeric integer value in [5, 300]. A pip is the minimum variation that can occur in the price of a currency. 
A stop-loss order is designed to limit an investor’s loss on a position. A take-profit order specifies the number of pips from the current price point to close out their current position for a profit. The trailing stop is a stop order that can be set at a defined percentage away from a security's current market price. A trailing stop for a long position would be set below the security’s current market price; for a short post, it would be set above the current price.  Hence , the second chromosome encodes three parameters: SL for the "stop-loss," TP for the "take-profit," and TS for the "trailing-stop."\\

\subsubsection{Chromosome related to triggering signals generation:}
For the MACD, the turnaround between the current bar and the previous one generates a buy signal. To do this, we include parameters; MACDpa and MACDpb, which will try to locate what point the signal is more reliable (Figure 2). The value is given in pips and between [0.100, -0.100]. 
 
\begin{figure}[h!]
\label{fig:2}
\includegraphics[width=\textwidth]{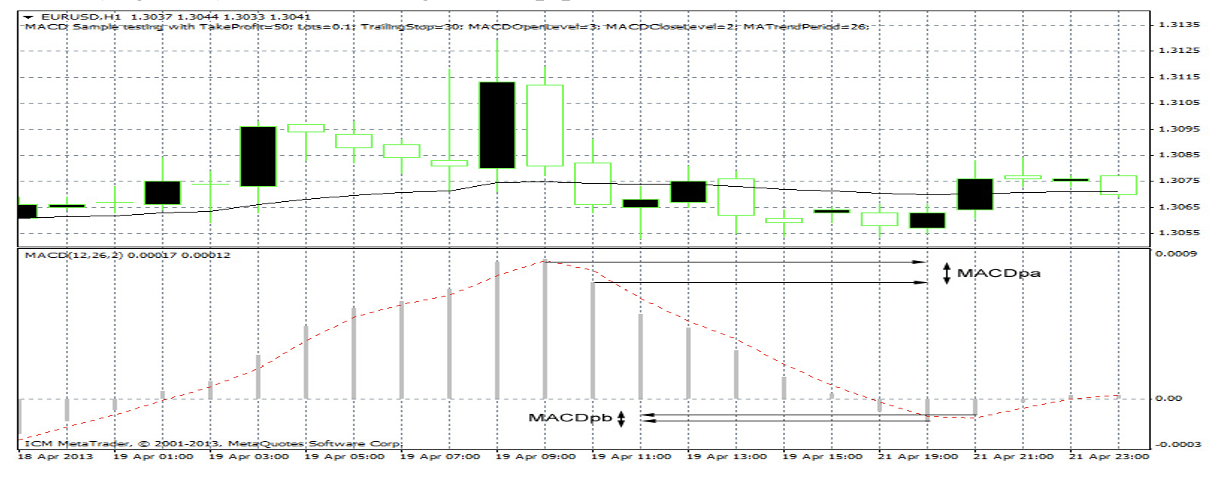}
\caption{MACD indicator parameters.}
\end{figure}
 
Figure 2. MACD indicator parameters.
The Stochastic (Figure 3) has two reference values , from which the overbought or oversold is generated. The parameters used are ESTpa, the higher reference level with values belonging to[50,100], and  ESTpb for the lower reference level, which is between [0.50]. 
 
\begin{figure}[!h]
\label{fig:3}
\includegraphics[width=0.7\textwidth]{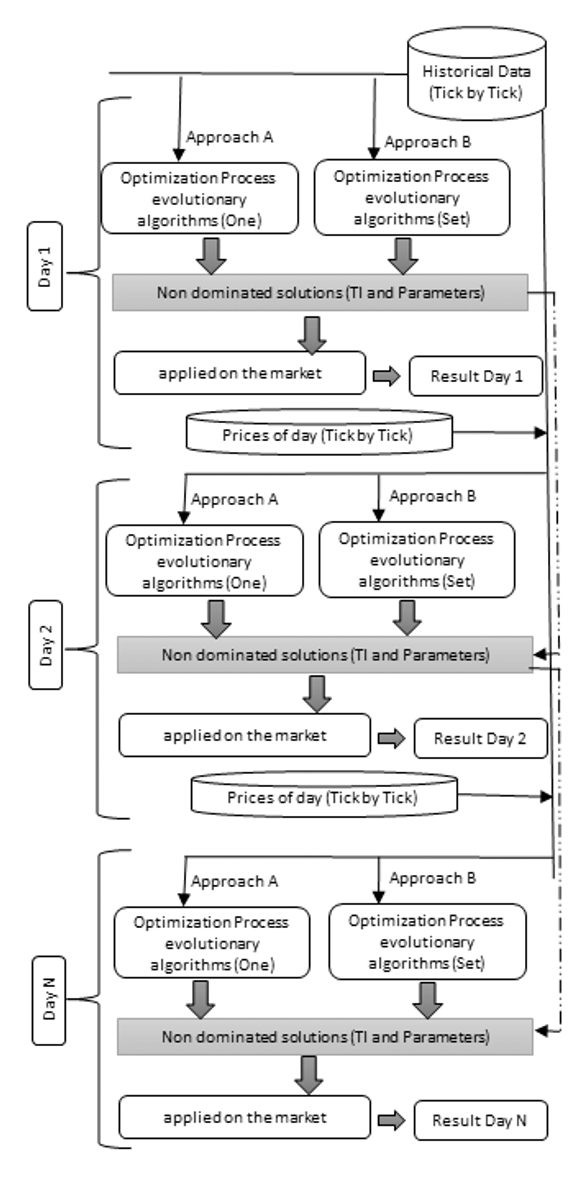}
\caption{Stochastic indicator}
\end{figure}

\subsection{Evaluation of the individuals}
\label{sec:evaluation}
The objective of the investment tool is to produce strategies to generate signals to buy and sell an asset in the market. The difference between the input and output prices establishes the operation's result. The operation can be positive or negative and may include the costs generated during the transactions. In this work, we propose tackling the problem from a multi-objective perspective. We will apply a set of fitness functions explained in this section. The fitness functions are based on a set of values of closed forms. Let us define them first. \\

\subsubsection{Definition 1:}
 The Holding Period Return (HPR) is the return obtained during the period it has been applied. For example, 1.01 and 0.99 refer to a gain and loss of 1\%, respectively.\\
 
\subsubsection{Definition 2: }The Terminal Wealth Relative (TWR) is the value obtained from multiplying all the HPRs. When HPRs are multiplied, a value that indicates the geometric mean (GM) is obtained as follows:
\begin{equation}
  \label{eq:5}
  GM = (TWR)^{1/N}
\end{equation}
being N  the number of operations.
The geometric mean value is proportional to TWR. Therefore, the higher the value of GM, the better the system performance.\\

\subsubsection{Definition 3:} Mathematical expectation (ME) is the amount expected to win or lose on average in each operation. The ME is a necessary but not sufficient condition for a profitable system since it does not take into account the number of operations. If a system has positive expectations and TWR is greater than one, then we have a profitable strategy, provided the same amount of capital is invested. To calculate the expectation of a system from its relationship percentage of success and gain/loss has the following expression:
\begin{equation}
\label{eq:6}
  ME = (1+B) * P-1  
\end{equation}

Where B is the Average (Profit / Loss) Ratio and P is the shooting percentage.\\

\subsubsection{Definition  4. The fundamental equation of trading.} 
GM, equation \ref{eq:5}, can also be estimated according to the arithmetic mean and standard deviation.
\begin{equation}
\label{eq:7}
    Geometric Mean = [Median^2-Std^2]^{N/2}
\end{equation}	(
GM and TWR  are related by the Nth root, as seen in definition 2. This allows estimating the TWR as:
\begin{equation}
    \label{eq:8}
    TWR = [Median^2-Std^2]^{N^2/2}
\end{equation}
This equation is called the "fundamental equation of trading". When the arithmetic mean is greater than one,  HPR will also be, and if the deviation is such that the term within the bracket is greater than one, then most operations will generate profits. The purpose is to maximize this aggregate function, divided into subcomponents to get different objective functions. 
 
\subsubsection{Objective Function 1: maximizing the arithmetic mean (maximize return)}
According to the above, it can perform the following definition.
RETURN OBTAINED BY THE STRATEGY FOR A OPERATION OX (a, b): Given an investment strategy consisting operate of a stock or index, between input and output signals generated by it and about a fixed set of parameters, in a given interval (a, b), is called return obtained by the position, the difference between the opening and closing price, subtracting the commissions.
\begin{equation}
    \label{eq:9}
    Ox= (Price_{open}-Price_{close}) – (interests)
\end{equation}
In this case, equation \ref{eq:9} coincides with the concept of HPR defined above. The only difference is that the HPRx is defined percentage and is calculated differently depending on whether a buy or a sale occurs.
\begin{equation}
    \label{eq:10}
    \begin{array}{cc}
       HPRx_{buy} = & 1+ \frac{Cf-Co}{Co} \\
        HPRx_{sell} =  & 1- \frac{Cf-Co}{Co} 
    \end{array}
\end{equation}	
In which Co is the price at the beginning of the operation plus his commission. Cf is the price at the end of the operation.
In a time interval, a set of positions can be opened. As above, we can make the following definition:
RETURN OBTAINED BY THE STRATEGY FOR A INTERVAL [a, b] (Bx(a, b)): Given an investment strategy, which operates about an asset or index, between the time instants a and b, according to the signals suggested by themselves and by a set of parameters fixed. It’s called the return obtained by the X strategy in the interval [a, b] to the sum of individual returns received by the operations in this interval.
\begin{equation}
\label{eq:11}
    Bx (a, b)=\sum_{i=a}^b O_{X_i} 	
\end{equation}

This definition coincides with that of TWR. The above equation (11) replaces the sum operator with a product operator because the HPRs are in percent.
\begin{equation}
\label{eq:12}
    TWR = \prod_{i=a}^b HPR_i
\end{equation}

With this term, we can calculate the geometric mean (5), which is used to calculate the performance of a system where profits are reinvested. A high value of this involves a higher system performance. 
To compare two strategies just have to calculate the geometric mean of both and know which one is greater. The geometric mean may also be calculated based on the arithmetic mean and standard deviation through "the fundamental equation of trading," explained at the beginning (7). 
Objective 1: Maximize the arithmetic mean, the TWR, and thus the return on investment.\\

\subsubsection{Objective Function 2: Minimize the risk of the strategy.}
Risk management is a combination of multiple interrelated parameters, among which are the following:
\begin{itemize}
    \item Contain the maximum number of consecutive losses ("drawdown")
	\item Controlling the dispersion of operations over the average. (Standard Deviation)
	\item Limit the amount of invested capital.
	\item 	Determine the limits on losses ("stop-loss") and profits ("take-profit").
\end{itemize}
A sound system should work with operations having a small dispersion from the mean. That is, it should check that the standard deviation is as small as possible. 
Thus, it is unnecessary to have a "stop-loss" and "take-profit" too large. Also, if we return to the initial concept, we see that when the arithmetic means are minimized, the geometric mean is maximized, thus, the ROI. 
Another risk factor is the permanence of market operations and the interest it generates. It is, therefore, essential to set a "take-profit" that allows collection profits and does not have an open position in a trading neutral. The only thing it does is consume resources and generate losses. Concerning ROI, it is referred to as a percentage of the initial investment.
In short, minimal deviation on average makes the operations work in a small range, limiting the operational time and the "drawdown." This maximizes the available capital for operations and minimizes risk.
Objective 2: Minimize the standard deviation (med) and adjust the "stop-loss" and the "take-profit" for maximizes the TWR.\\

\subsubsection{Objective Function 3: Maximize the number of operations.}
The investor always tries to reduce overall investment risk by diversifying her portfolio. This action allows you to find synergies that accelerate capital growth. When operating with a positive expectation system, further operations for the same period help maximize this increase and, therefore, its diversification. A diversified portfolio is a set of values that operate in conjunction to obtain a positive expectation. Correlation management is complex since when the market has a steep trend, all values are correlated, and gains or losses are widespread .
In the developed system, the securities portfolio consists of all possible solutions of the Pareto front in full. Each solution maximizes the number of operations you can get.  The system works with the same asset and currencies but can find different solutions. This means that in the same period, the solutions found by the system execute orders for buying and selling simultaneously. This produces the diversification of the complete portfolio.
Another factor to consider is the ROI. One strategy with a very optimistic expectation but only operates once offers worse performance than a strategy with a positive expectation and operates many times. An investor always seeks the best possible return on their investment, which requires increasing the number of operations.
Objective 3: maximize the number of operations N, maximize the diversification and performance.

\section{Experimental Results}
\subsection{experimental Framework}
We implemented a real-time software tool for testing the ideas presented in previous sections.  It has been developed under the object-oriented paradigm with two programming languages: C ++ and Java. The heuristic algorithms were implemented using the jMetal 4.0 © [19] library. jMetal is an object-oriented framework for developing, experimenting and studying metaheuristics for solving multi-objective optimization problems. Metatrader 4 © software is utilized for data acquisition and real-time operation. MetaTrader is an online trading platform that provides brokerage services for real-time market investment. In addition, Metratader can be used to perform technical analysis on specific temporary intervals. It also allows using different time investment windows, ranging from minutes to months. Our Software architecture permits the application of the heuristic algorithms of jMetal on the whole set of assets available at Metatrader. 

Our System Architecture, which is represented in Figure 4, consists mainly of six modules (two of them are databases structures): 
\begin{itemize}
\item The data collector obtains values of assets in real-time and stores the information. This module is always running.
\item The information manager, The Information Manager is responsible for managing the system's response from data provided by the experiments manager and the active experts. It determines the experts who should be engaged in real-time.
\item The manager of experiments: The Experiments Manager handles the parameters, such as the number of evaluations, the temporary period to analyze, the Evolutionary Algorithm used, its probability values of mutation and crossover, etc. This can be used for conducting experiments or for operation in real-time. The only thing that changes is the dataset used.
\item The expert: The Expert (business solution) is responsible for making the buy and sell and works directly on the trading platform. The data needed to operate are generated in real-time and are stored in the operational database. It can operate concurrently with all solutions along the Pareto frontier. 
\item A database storing the historical data
\item A database updated in real-time containing the most recent data and the non-dominated solutions.
\end{itemize}

\begin{figure}[!h]
\label{fig:4}
\includegraphics[width=0.75\textwidth]{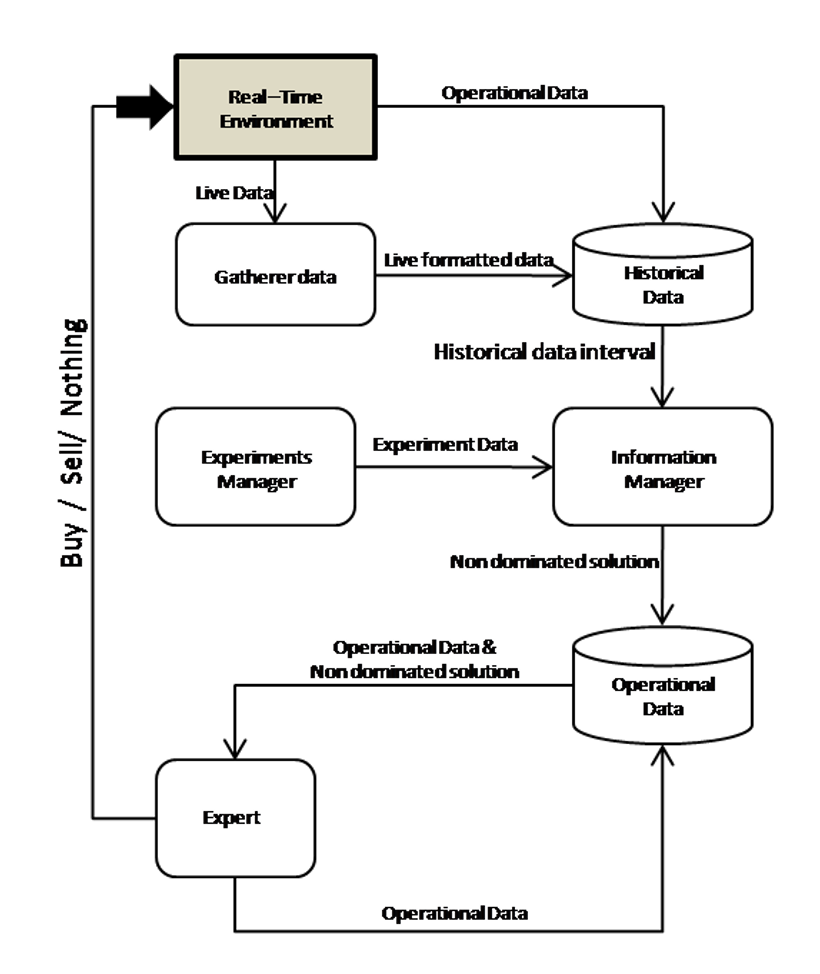}
\caption{System Architecture diagram.}
\end{figure}

\subsection{Experiments}
This study aims to compare the optimization from both approaches multi-objective. According to several authors [35], the best strategies for "trading" are generated by combining two indicators, a follower of trends and an oscillator. Following this recommendation has been prepared the method used in this work has. The selected indicators were MACD and stochastic \cite{elder1993trading}. The first will indicate the so-called "surge," the system's tendency, while the second detects the best operating options following the previous trend, also called "the wave." Both indicators are explained in the section on technical indicators (section 3.2).
This paper has selected a basic “trading strategy” for conducting experiments. This has been optimized using evolutionary algorithms to improve the results obtained by other classical approaches.  Finding the best trading strategy for this market is beyond the scope of this work.  
The set of experiments is as follows:
\begin{enumerate}
    \item Experiment A1, A2, and A3: Considering approach A, with the data sets 1, 2, and 3, respectively.
    \item 	Experiment B1, B2, and B3: Considering approach B, with the data sets 1, 2, and 3, respectively.
\end{enumerate}
For this study, we used the EUR / USD currency market dataset. The data sets were selected because they represent a sufficiently long period and constitute different market trends. Table 1 shows a summary.
\begin{table}[]
\begin{tabular}{|l|l|l|l|}
\hline
Set & Interval                & Amount of data & Trends  \\ \hline
1   & 2004/03/29 - 2004/05/02 & 800.000        & Neutral \\ \hline
2   & 2006/03/27 – 2006/05/30 & 660.000        & bullish \\ \hline
3   & 2008/03/31 – 2008/06/30 & 675.000        & both    \\ \hline
\end{tabular}
\caption{Data sets selected for experiments}
\label{tab:table1}
\end{table}

Data have been obtained from the website http://ratedata.gaincapital.com. These have a format bid/ask and are "tick by tick."
\subsection{Results and discussion}
First, we want to study the benefits of each proposed approach. In both cases, the algorithms used are PESA-II, PAES, NSGA-II, SPEA-II, and MoCell. "Approach A" comprises non-dominated solutions of the Pareto front of the set of all selected evolutionary algorithms. While "approach B" will be formed only of those non-dominated solutions belonging to the Pareto front of any of the above algorithms. In both cases, all will be used all solutions to operate. Figures 5, 6, and 7 show the performance obtained for the compared methods and datasets.
In all cases, you can appreciate the superiority of the approach proposed in this paper. The results show that the best values are obtained by method B compared to any evolutionary algorithms in approach A. Therefore, it is concluded that the proposed scheme in Figure 4 represents a good model that can be applied to analyze investment strategies stock market.
\begin{figure}[ht!]
\label{fig:5}
\includegraphics[width=\textwidth]{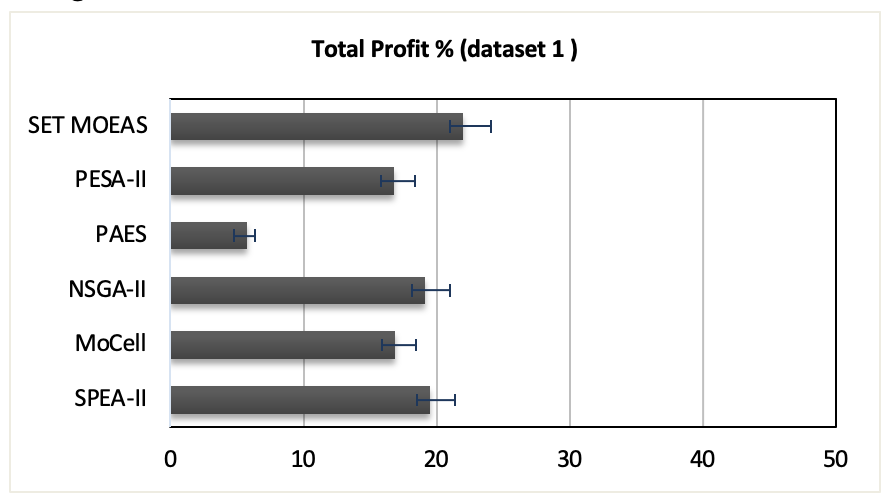}
\caption{Total profits of the different approaches, in percent, for dataset 1.}
\end{figure}
 
\begin{figure}[ht!]
\label{fig:6}
\includegraphics[width=\textwidth]{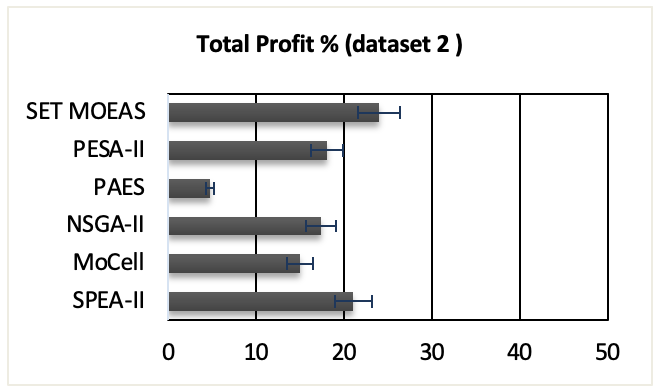}
\caption{Total profits of the different approaches, in percent, for dataset 2.}
\end{figure}

\begin{figure}[ht!]
\label{fig:7}
\includegraphics[width=\textwidth]{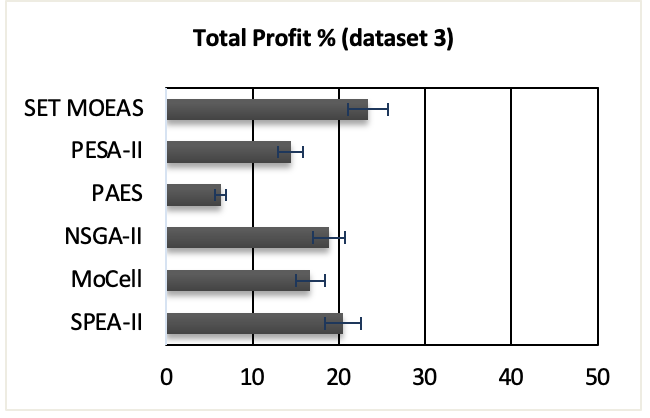}
\caption{Total profits of the different approaches, in percent, for dataset 3}
\end{figure}

In each new iteration, new non-dominated solutions are generated. Other non-dominated solutions replace some before performing any operation. Figures 8, 9, and 10 show the number of various solutions developed in a complete experiment and which it operates. 

\begin{figure}[ht!]
\label{fig:8}
\includegraphics[width=\textwidth]{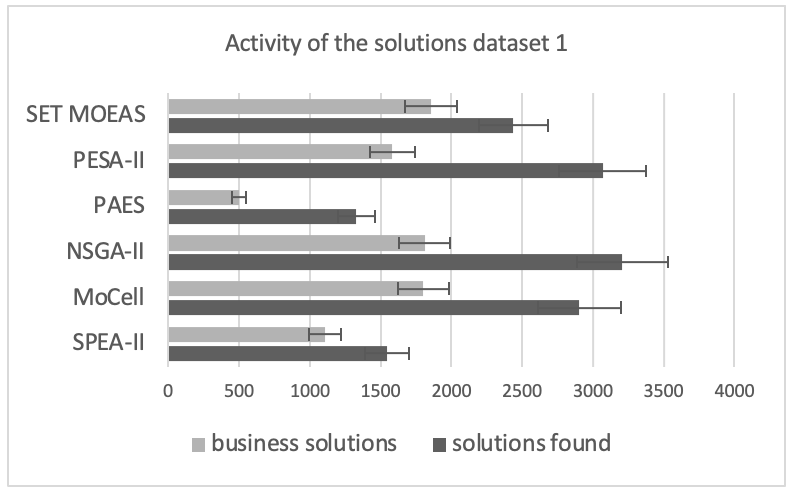}
\caption{The activity of the solutions found compared with those that perform some operation (Business Solutions) for the data set 1.}
\end{figure}

\begin{figure}[ht!]
\label{fig:9}
\includegraphics[width=\textwidth]{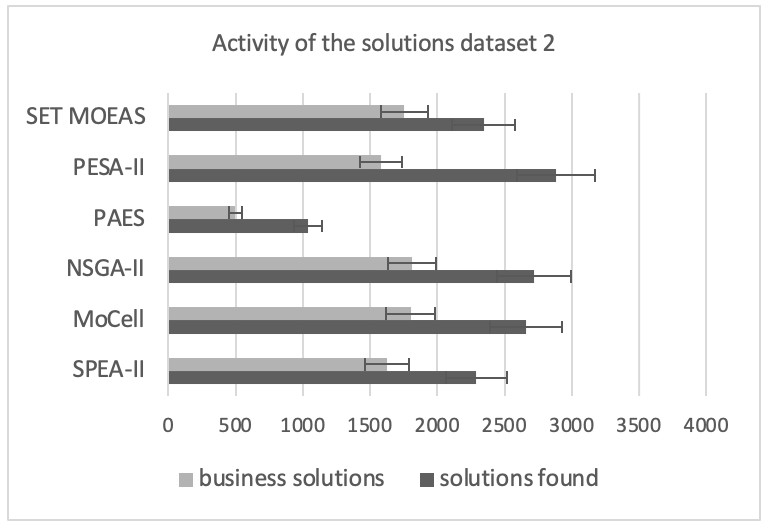}
\caption{The activity of the solutions found compared with the solutions that perform some operation (Business Solutions) for the data set 2.}
\end{figure}

\begin{figure}[ht!]
\label{fig:10}
\includegraphics[width=\textwidth]{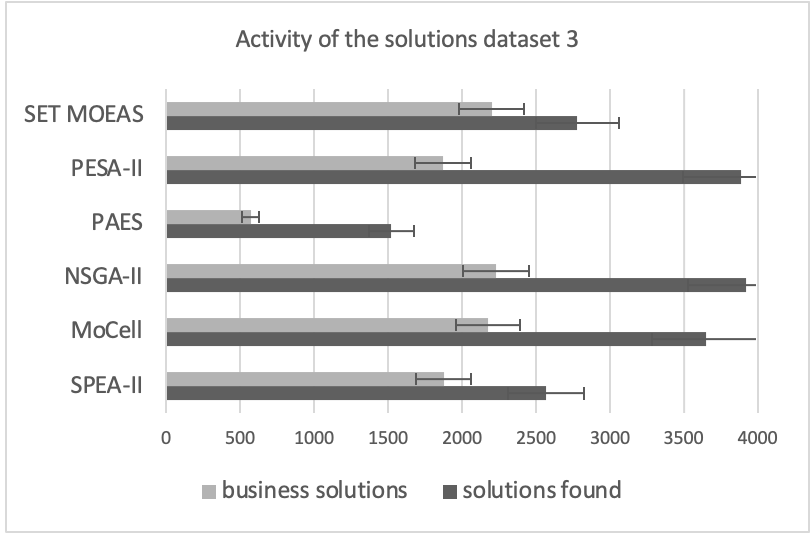}
\caption{The activity of the solutions found compared with the solutions that perform some operation (Business Solutions) for the data set 3.}
\end{figure}

Also, we see that for approach B, the number of "business operations" is far superior to the rest. This is a feature which that makes the proposed method either more optimal than others. We observe this fact in more detail in Figure 11. Moreover, the influence of different data sets in terms of variation of the solutions obtained is irrelevant.

\begin{figure}[h!]
\label{fig:11}
\includegraphics[width=\textwidth]{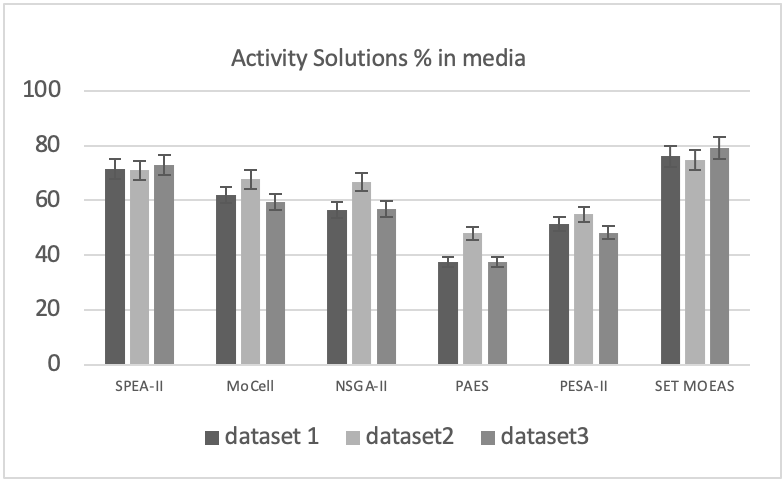}
\caption{Summary of "Business Solutions" activity for the entire data set.}
\end{figure}
 
In Figure 12, you can see the average profit that has a solution for every operation of purchase/sale made while it is active. There are variations in the results obtained by different evolutionary algorithms. However, no significant between those obtained by the two methods proposed. This is normal since, in any case, the proposed method does not involve an improvement in the performance of different evolutionary algorithms treated.

\begin{figure}[h!]
\label{fig:12}
\includegraphics[width=\textwidth]{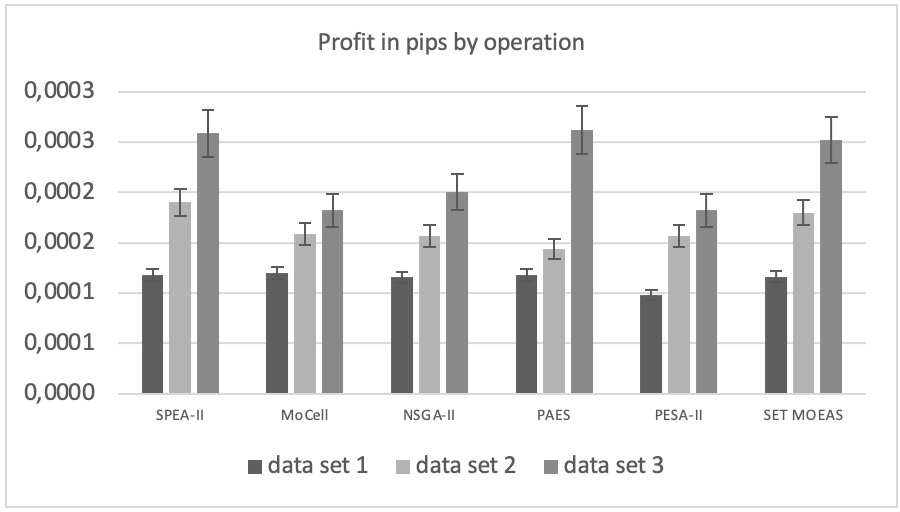}
\caption{Profit in pips for each operation performed for all data sets.}
\end{figure}

This section compares the benefit obtained by the technique called "Approach B" with others. Most of the works in the area only operate with a single solution. This solution is sometimes maintained for the entire period from a previous optimization \cite{lohpetch2011multiobjective}. In other cases, it changes with the arrival of new data, but ultimately only one solution operates despite using different targets to obtain the final solutions \cite{bodas2009multiobjective}. This selection can depend on various factors but typically is determined by the gain. This line is covered by what we call "Approach A." 
Moreover, the selected strategy is compared with the "Buy \& Hold" strategy. This strategy is widespread to evaluate investment strategies \cite{chen1997toward}. This strategy involves buying a security and keeps for a long time. The benefit of the Buy \& Hold strategy is obtained by subtracting the initial value from the final value. Possible operations are two since the investor can bet that the value will increase or decrease. In this study, 100 experiments are used to simulate each type of strategy. The following figure shows the results obtained (Figure 13). 

\begin{figure}[h!]
\label{fig:13}
\includegraphics[width=\textwidth]{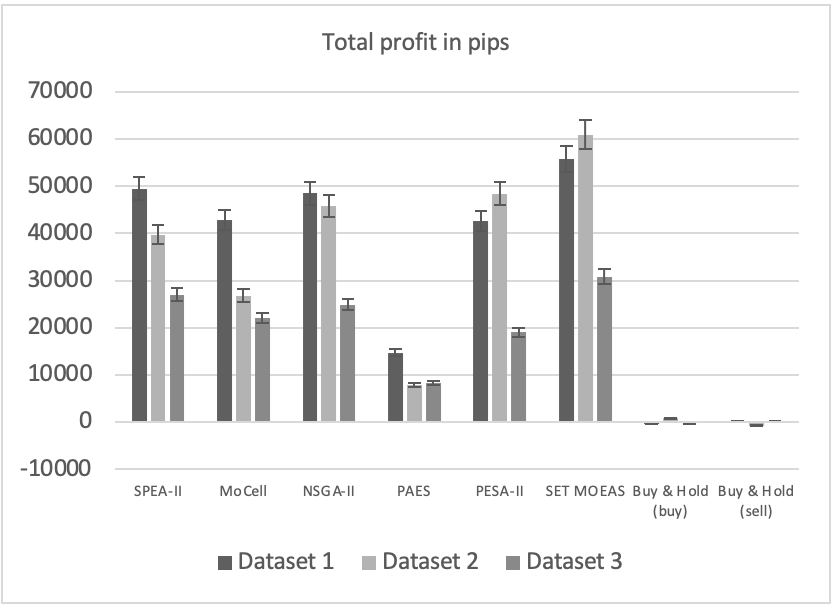}
\caption{Performance of our method (Approach B) in comparison with the "Buy \& Hold" strategy and individuals approach (Approach A).}
\end{figure}

The approximation B improves the other results. Consequently, we conclude that the proposed approach is a valid research strategy.

\section{Conclussions and Future work}
A tool for real-time operations in the foreign exchange market has been developed. It is based on optimizing a set of parameters via a multi-objective evolutionary algorithm. These are related mainly to three areas: the stock technical indicators, other specific market indicators, such as stop-loss, take-profit, and trailing-stop, and, finally, the corresponding time windows in which the operation takes place. The system offers high scalability to other potential financial markets. It is also entirely customizable for any type of indicator or the inclusion of new parameters.\\

The system generates a set of experts that operates autonomously in the market. They are continually evaluated to see if others get better performance for the current data set, and if it occurs, they are replaced automatically. This set consists of the entire Pareto front of solutions of a group of MOEA’s; they usually make up between 80 and 120 different solutions. Sometimes they can be used as filters to reduce the number of experts and to limit the number of operations.\\

Evaluation of different Evolutionary Algorithms has permitted observing how, in general, the optimization process works best. For the selected market in this work, the set of MOEA’s has reached the best yield. Further adjustment of this one has resulted in a considerable increase in its ultimate performance.
The tool generates different operational conditions for each expert (solution) depending on the degree of success obtained in the above range. This is very interesting; as an expert who has generated a very poor profit or even losses, it is not eliminated, but the amount of operational risk can be reduced, even being able to be zero in some cases. The decision to replace an expert is solely attributable to being dominated by another. This behavior can limit the risk locally and globally for all transactions. This is one of the main objectives of any system of trading.\\

It has been shown that the maximization of the number of transactions improves profits. The total cost of transactions increases, but the gain far outweighs this cost. Also, when performing many operations resolved briefly, the costs associated with interests are reduced to virtually zero. It was also noted that the system tended to obtain the benefits quickly.
The evolutionary approach is very robust since all non-dominated solutions obtained from Pareto fronts are similar in quality. The experience acquired with fine-tuning the algorithm will be helpful with other indicators.\\

Future versions will include more indicators allowing the user to choose the indicators he wants to operate. The system should also incorporate more complex models of trading costs. It would be helpful to research new methods for estimating the profitability reached by the strategy and applying it to other indexes and markets. Finally, a complex risk management system could fit the profile of particular investors, providing estimations of potential profit increases assuming these risks.\\

%
% ---- Bibliography ----
%
% BibTeX users should specify bibliography style 'splncs04'.
% References will then be sorted and formatted in the correct style.
%
\bibliographystyle{splncs04}
\bibliography{economy}
\end{document}